\documentclass[letterpaper, 10 pt, conference]{ieeeconf}  

\usepackage{times}

\usepackage{multicol}
\usepackage{multirow}
\usepackage{hyperref}

\usepackage{amssymb}
\usepackage{amsmath}
\usepackage{textcomp}
\usepackage{graphicx}
\usepackage{dblfloatfix}
\usepackage{cuted}    
\usepackage{capt-of}

\usepackage{booktabs}
\usepackage[table]{xcolor}
\usepackage{placeins}
\usepackage{supertabular}   

\title{\LARGE \bf
Humanoid Whole-Body Badminton via an \\Annealed Reinforcement Learning Curriculum
}

\author{Chenhao Liu*, Leyun Jiang, Ningyuan Tian, Yibo Wang, Jincheng Fu, Kairan Yao, Xiaoyu Ren}

\begin{document}

\maketitle
\thispagestyle{empty}
\pagestyle{empty}

\begin{strip}
\vspace*{-3\baselineskip}
  \centering
  \includegraphics[width=\textwidth,height=0.35\textheight]{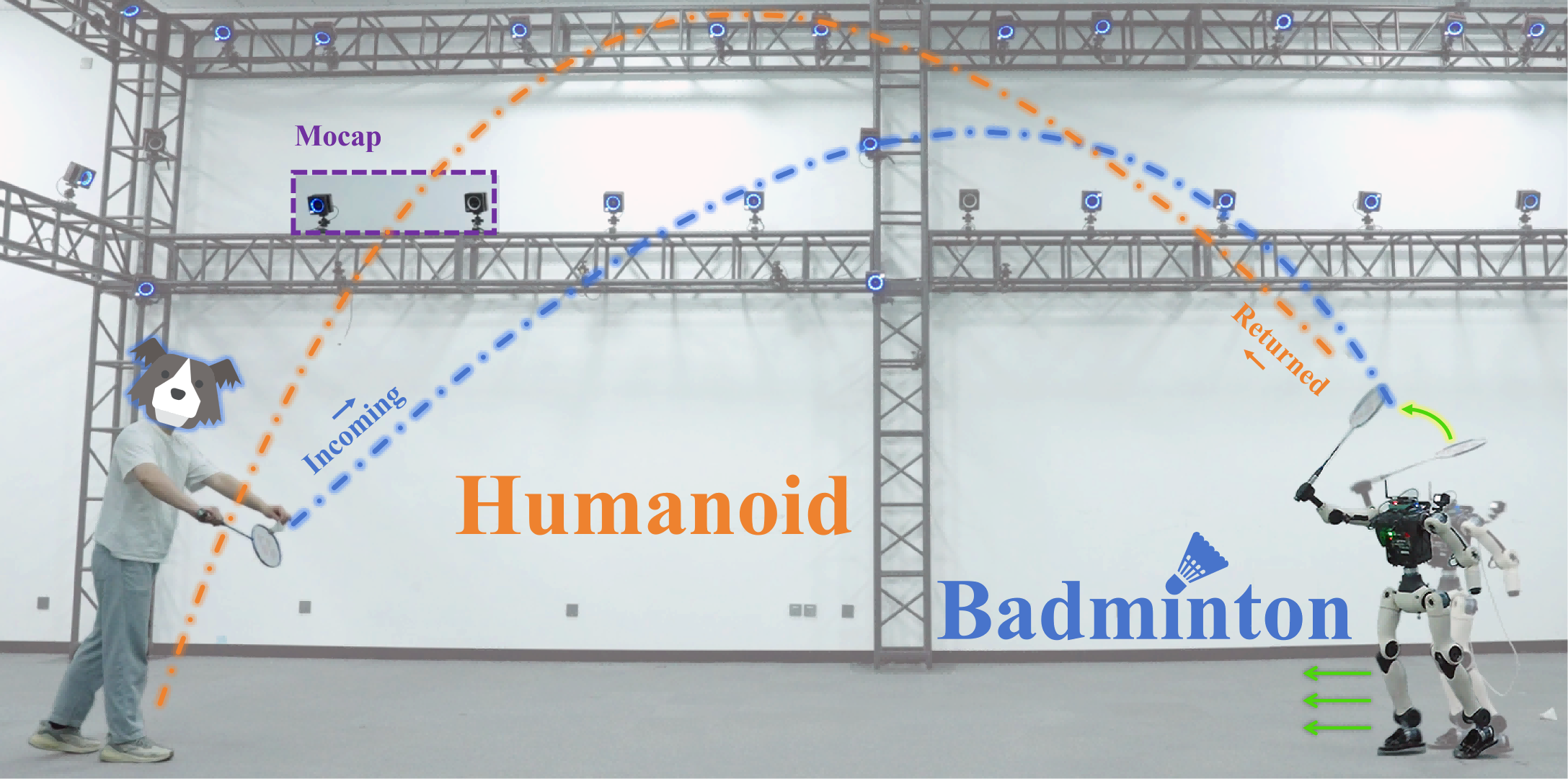}
  \captionof{figure}{\textbf{Real-world humanoid badminton.} A fully autonomous humanoid rallies with a human player in a motion capture arena; overlaid arcs show an incoming (blue) and returned (orange) trajectory. Project Page: \href{https://humanoid-badminton.github.io/Humanoid-Whole-Body-Badminton-via-Multi-Stage-Reinforcement-Learning/}{humanoid-badminton.github.io}.}
  \label{fig:figure1}
\end{strip}



\vspace*{-1.5\baselineskip}
\begin{abstract}
Humanoid robots have demonstrated strong capabilities for interacting with static scenes across locomotion and manipulation, yet dynamic real-world interactions remain challenging. As a step toward fast-moving object interactions, we present an RL training pipeline that yields a unified whole-body controller for humanoid badminton, coordinating footwork and striking without motion priors or expert demonstrations. In badminton, locomotion and striking are tightly coupled, making the final hitting objective difficult to optimize directly due to sparse rewards and conflicting gradients. We address this with an annealed curriculum that first stabilizes learning through auxiliary locomotion objectives, then progressively removes them to focus optimization on the final hitting objective. For deployment, we use an Extended Kalman Filter (EKF) to estimate and predict shuttlecock trajectories for target striking, and also develop a prediction-free variant that removes the EKF and explicit prediction. We validate the framework in simulation and on hardware. In simulation, two robots sustain a rally of 21 consecutive hits. In real-world tests with both machine-fed shuttles and human-robot rallies, the robot achieves outgoing shuttle speeds up to 19.1~m/s. Moreover, the prediction-free variant attains comparable performance to the EKF-based policy. Overall, our approach enables dynamic yet precise goal striking in humanoid badminton and suggests a path toward more dynamics-critical whole-body interaction tasks.
\end{abstract}

\section{Introduction}


Humanoid platforms have been proposed as general-purpose embodied agents for human-compatible skills \cite{radosavovic2024real, truong2025beyondmimic, qiu2025humanoid, zhuang2024humanoid, he2025hover}. Despite rapid progress in locomotion and motion imitation, agile, contact-rich interactions with fast-moving objects under tight reaction windows remain underexplored. Badminton is an ideal testbed: returns require sub-second perception–action loops, precisely timed and oriented racket–shuttle contacts within a large 3D interception volume, and whole-body coordination that blends rapid arm swings with stable and agile leg motions.

Compared with several recent works on robotic table tennis \cite{dambrosio2025achieving, d2023robotic, su2025hitter, hu2025towards}, badminton aggravates several difficulties. Aerodynamics are highly uncertain: strong drag and a special “flip” regime make the pre-turning trajectory hard to model; as a result, the seemingly longer flight still yields less usable decision time. Although the racket face is larger, valid hits occur only in brief windows when position, velocity, and orientation align; swing amplitudes are also much larger, injecting angular momentum that challenges whole-body balance. Critically, footwork and striking must co-evolve: lower-body motion does not merely reposition the base but co-determines hitting accuracy. Our system reaches swing speeds about 5 m/s and produces outgoing shuttle speeds up to 19.1 m/s. The serve-to-hit window is typically less than 1.0 s; the first 0.36 s is consumed by trajectory prediction, leaving less than 0.7 s for the controller to react.


Related efforts highlight both promise and gaps. \textit{HITTER} demonstrated humanoid table tennis via a hierarchical planner coupled to a learned whole-body controller and achieved long rallies \cite{su2025hitter}. Our setting differs in key ways: (i) we do not use reference motions to shape striking; instead, the policy discovers energy-efficient swings purely via reinforcement learning, simplifying implementation, and avoiding excessive style constraints; (ii) the use of “virtual hit plane” reduces striking to a largely 2D problem, whereas badminton demands orientation-aware contacts throughout a 3D space;  (iii) badminton’s large-amplitude strokes induce significant whole-body disturbance, requiring balance recovery that table-tennis-scale motions seldom provoke; and (iv) we do not explicitly command target base positions; instead, the policy only receives the desired interception target and must coordinate the whole-body motion.  As for badminton, recent work on a legged-manipulator learns coordinated skills with onboard vision \cite{scirobotics.adu3922}; however, humanoid hardware deployment has remained elusive, and their simulated humanoid badminton policies did not exhibit badminton-style footwork, suggesting missing training signals for lower-body coordination. These gaps motivate a training pipeline that explicitly fosters footwork–strike synergy on humanoids and scales to real hardware.


This paper presents, to our knowledge, the first real-world humanoid robot that plays badminton. We train a unified whole-body controller with an annealed RL curriculum designed to resolve the optimization conflict between footwork and striking while preserving a single deployable policy. Rather than primarily increasing task difficulty, our curriculum anneals away auxiliary shaping objectives, making the final objective increasingly dominant. Concretely, we instantiate this curriculum as a three-stage training procedure—footwork acquisition, precision-guided swing generation, and task-focused refinement. We adopt the shuttlecock simulation of \cite{scirobotics.adu3922} to generate a corpus of about 20k flight trajectories for training. At each rollout, the policy receives a compact target tuple (hit time, intercept position, and target racket orientation) sampled from this pre-generated corpus, with intercepts drawn from a 1.5--1.6~m height band suited to the humanoid's scale. 

The policy observes hitting target information as well as proprioception and inferred actions for 21 joints on a 1.28 m humanoid. In deployment, the hitting target is obtained online from an EKF that predicts the shuttlecock trajectory. We leverage a low-level PD controller that runs at 500 Hz, with policy inference at 50 Hz to derive the desired joint positions. The policy is trained fully in simulation with moderate domain randomization, and is deployed in a zero-shot manner without system identification. We conduct real-world experiments in a motion capture (MoCap) arena.

Beyond our EKF-based pipeline, we also explore a non-hierarchical, reactive prediction-free variant that removes any explicit shuttle predictor. In contrast to recent humanoid table tennis systems where end-to-end RL is combined with an auxiliary learnable predictor that outputs hitting targets for a downstream controller \cite{hu2025towards}, our variant does not introduce a separate prediction module. During training, the actor only receives the instantaneous shuttle position plus five history frames to infer hit timing and target pose, while the critic retains privileged target to better learn the value function. This more end-to-end control strategy is conceptually closer to human-like rolling anticipation and simplifies deployment by avoiding additional predictor or planner tuning, and naturally supports aerodynamic randomization. We report comparison results in both simulation and hardware evaluation.

Both simulation and real-world results indicate that our humanoid can execute large-amplitude swings while maintaining balance, react within sub-second windows, and intercept fast incoming shuttles fully autonomously. The key technical contributions are summarized as follows:

\begin{itemize}
    \item  First real-world humanoid badminton: unified whole-body control for returning machine-served shuttles and demonstrating human–robot rallies with a 21-DoF humanoid.

    \item  Annealed curriculum for footwork–strike coordination: auxiliary locomotion objectives stabilize early training and are progressively removed to prioritize precise striking through three stages: footwork acquisition, swing generation, and task-focused refinement.

    \item  A prediction-free variant: the policy infers striking timing and targets directly from short-horizon shuttle observations, improving robustness to aerodynamic variation and removing the need for explicit trajectory prediction.
\end{itemize}

\begin{figure*}[!t]
  \centering
  \includegraphics[width=\textwidth, height=0.35\textheight]{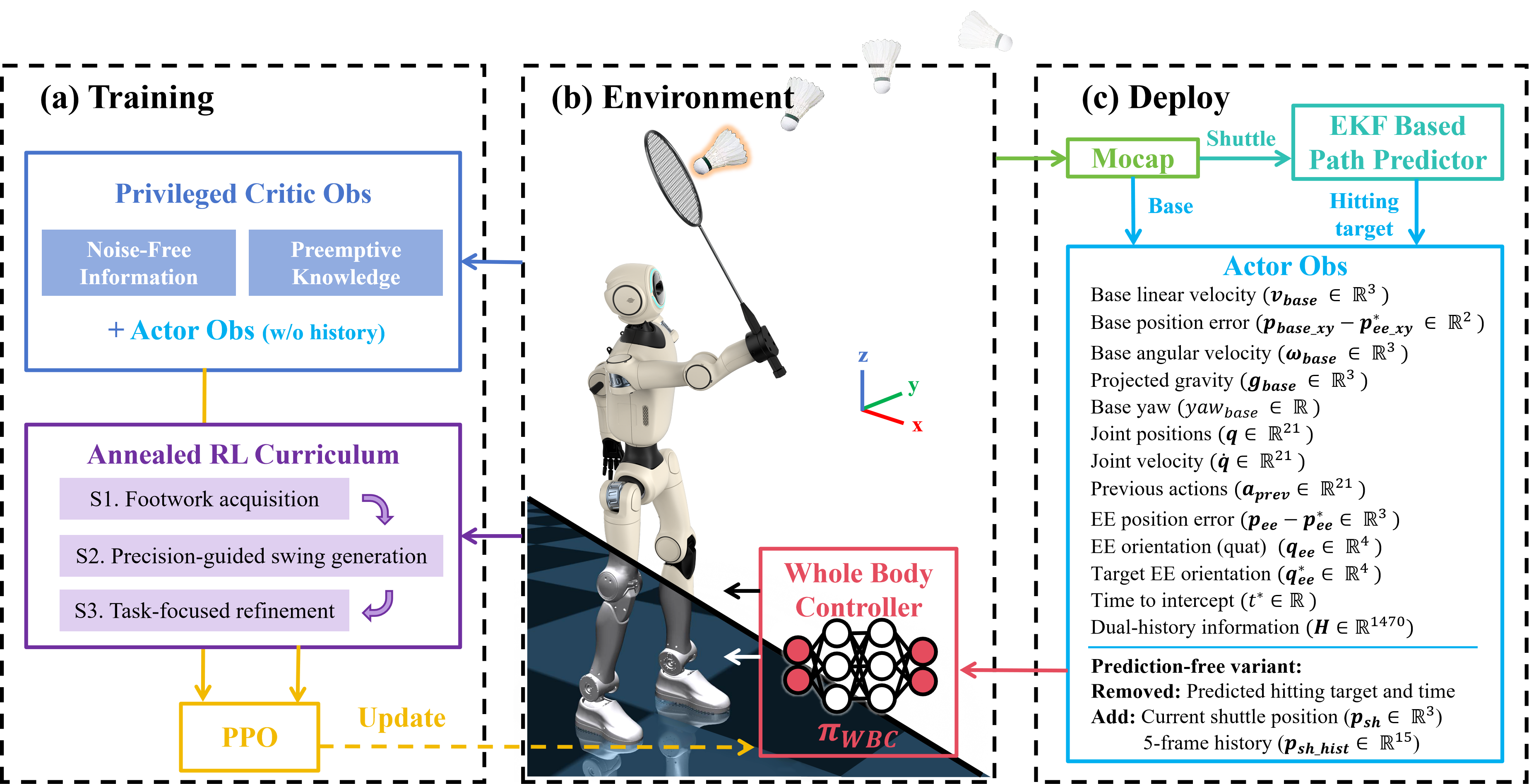} 
  \caption{\textbf{System overview.}
\textbf{(a) Training:} PPO learns a single policy $\pi_{\mathrm{WBC}}$ using Privileged Critic Obs and Actor Obs through a three-stage annealed curriculum. All observations and rewards in (a) come from the simulation.
\textbf{(b) Environment:} The humanoid is initialized above the origin and faces the $+x$ direction. A 3D-printed mount attaches the racket orthogonally to the forearm.
\textbf{(c) Deploy:} At runtime, MoCap directly provides the base state, while an EKF with a path predictor estimates the shuttle trajectory to produce the hitting target $\{\boldsymbol{p}^*_{\mathrm{ee}}, \boldsymbol{q}^*_{\mathrm{ee}}, t^*\}$. For the prediction-free variant: We discard the EKF-based prediction and replace the hitting target with current shuttle position $\boldsymbol{p}_{\mathrm{sh}}$ and a 5-frame history $\boldsymbol{p}_{\mathrm{sh\_hist}}$, keeping all other components unchanged.
These information, together with proprioception and dual-history features, forms the complete Actor Obs. $\pi_{\mathrm{WBC}}$ consumes the Actor Obs and outputs whole-body actions executed by a low-level PD controller.}
  \label{fig:figure2}
\end{figure*}

\section{Related Work}

\subsection{Whole-Body Loco-manipulation}
Combining manipulation with mobility is essential for many robotic applications. A common approach is to decouple base motion and end-effector tracking to reduce task complexity. For example, an MPC controller tracks the desired end-effector position of an arm mounted on a quadruped, while a standalone RL controller handles locomotion \cite{9684679}. Other works decompose dynamic catching into separate base, arm, and hand policies \cite{11127596}, or assign upper- and lower-body control to different agents to improve end-effector tracking accuracy during locomotion \cite{li2025learning}. In contrast, there have been many efforts to unify lower and upper body control. Several works use multi-critic architectures to ease the optimization of coordinated upper- and lower-body behavior \cite{fu2023deep, vijayan2025multi, wang2024arm}, while others introduce physical-feasibility rewards to guide unified policy learning \cite{hou2025efficient}. For dynamic whole-body tasks such as tossing, performance hinges on coordinated whole-body impulse generation and timing, with the legs contributing to power transmission rather than merely locomotion; therefore, unified whole-body controllers have proved effective in these domains \cite{ha2024umi, ma2025learning}.

\subsection{Robotic Racket Sports}
Robotic racket sports have become a key benchmark for dynamic whole-body loco-manipulation \cite{su2025hitter, hu2025towards, scirobotics.adu3922, wang2025integrating, 9797506, zaidi2023athletic}. The tight coupling between perception, fast interception, and precise end-effector control makes these tasks particularly challenging. One system achieves human-level table tennis through a highly modular and hierarchical architecture combining learned and carefully engineered components \cite{dambrosio2025achieving}. However, it requires substantial manual design, module-wise tuning, and large-scale real-world data collection. Subsequent works extend this direction to humanoid table tennis. \textit{HITTER} combines an RL-based whole-body controller with a model-based planner for ball trajectory prediction and racket target planning, enabling human–robot rallies, but relies on expert demonstrations for reference motions \cite{su2025hitter}. Another study jointly trains an auxiliary module during RL to predict ball trajectories and propose racket targets \cite{hu2025towards}. In both systems, base motion is separately commanded rather than jointly optimized with upper-body striking under a unified task objective. A perception-informed whole-body policy has also demonstrated impressive shuttlecock striking on a quadrupedal platform \cite{scirobotics.adu3922}. However, quadruped morphology substantially simplifies balance compared with a humanoid with a narrow support polygon and high center of mass, making the coordination of balance and large-amplitude racket swings considerably less challenging. In this work, we instead focus on coordinated badminton skill learning on a humanoid platform. We train a single whole-body controller through an annealed RL curriculum, without expert demonstrations or separate base commands, and evaluate it in simulation and real-world experiments.

\section{System Overview}

Experiments use a 1.28 m, 30 kg, 21-DoF humanoid with IMU and joint encoders, comprising 6 DoF per leg, 1 DoF at the waist, and 4 DoF per arm. A standard 100 g full-size badminton racket is rigidly clamped to the distal forearm, as shown in Fig. \ref{fig:figure2}(b). In the MoCap arena, markers are placed on the robot base to provide base position, orientation, and linear velocity, while the shuttle carries reflective tape on its tip and provides position only. The target-known pipeline relies on an EKF to estimate and predict the shuttle state, whereas the prediction-free variant removes explicit prediction and instead conditions the policy on a short history of observed shuttle positions. The MoCap system runs at 210 Hz, the policy at 50 Hz, and a joint-space PD controller at 500 Hz.

\section{Learning Coordinated Badminton Policy}

\subsection{RL-based Dynamic Whole-Body Controller}
\subsubsection{MDP Formulation and Annealed Curriculum}

We cast policy learning as a partially observable Markov Decision Process (MDP) \(\mathcal{M}=(\mathcal{S},\mathcal{O},\mathcal{A},\mathcal{P},\mathcal{R})\).
At time \(t\), the (unobserved) state is \(s_t\in\mathcal{S}\); the agent receives an observation \(o_t\in\mathcal{O}\), applies action \(a_t\in\mathcal{A}\), transitions with \(s_{t+1}\sim\mathcal{P}(\cdot\mid s_t,a_t)\), and accrues reward \(r_t=\mathcal{R}(s_t,a_t)\). We learn a parametric policy \(\pi_\theta(a_t\mid o_t)\) that maximizes \(J(\theta)=\mathbb{E}[\sum_t \gamma^t r_t]\).

\textit{Observation space.} The actor’s basic observation is 87-D, concatenating proprioception information and external sensing from MoCap. On top of this 87-D vector, we stack short/long history of all joint positions, velocities, and actions \cite{li2025reinforcement}, adding another 1470 features to mitigate partial observability. The critic observation has 98 dimensions in total, which include the privileged information used to enhance actor version observations, such as noise-free base and joint states as well as racket speed which may be difficult to acquire in the real-world. Others are classified as preemptive knowledge used to make MDP well-posed for accurate value learning. Detailed information is listed in Fig. \ref{fig:figure2}(a), (c).

\textit{Action space.} The policy outputs 21 joint position targets for all DoF, scaled by a unified action scale of 0.25 and tracked by the low-level PD controller.

\textit{Episode Settings.} Each episode contains six swing targets, following the configuration used by \cite{scirobotics.adu3922}. This design enables follow-through behavior learning but makes the value function depend on the number of hits remaining. We therefore adopt an asymmetric actor–critic \cite{Pinto-RSS-18}, the critic receives additional preemptive information to stabilize value estimation, while the actor only uses deployable observations. Concretely, these are two subsequent hitting times, next hitting target, and the number of targets remaining .


\textit{Training Settings.} We train with PPO \cite{schulman2017proximal} on a single Nvidia RTX 4090 using 4096 parallel environments in IsaacGym \cite{makoviychuk2021isaac}. Both actor and critic are MLPs with hidden layers 512–256–128 and ELU activations.

\textit{Annealed curriculum.} As shown in Fig. \ref{fig:figure2}(a), we instantiate the annealed curriculum as a three-stage training schedule, where auxiliary locomotion shaping is gradually reduced while the observation and action spaces remain unchanged. Training in each stage continues from the previous checkpoint once the current stage's primary objective has converged. The whole process takes 12.4 h wall-clock time, after which the resulting policy is ready for deployment. The three stages are organized as follows:

\begin{itemize}
    \item S1 — Footwork acquisition toward sampled hit regions: the policy learns to approach target regions with reasonable lower-limb gait while maintaining a stable, forward-facing posture as it traverses between six sampled hit locations within an episode.

    \item S2 — Precision-guided swing generation: building on S1, we introduce a sparse hit reward active only at the scheduled hit instant \(t^*\) to enforce timing. We start with loose pose accuracy requirement to allow a full swing to emerge, then progressively tighten position and orientation tolerances. We also add light swing-style regularization and strengthen energy, torque, and collision constraints for efficient and stable hits.

    \item S3 — Task-focused refinement: starting from S2, we anneal away the target-approach reward and multiple gait-shaping regularizers, including foot distance, step/contact symmetry, and step height, to reduce gradient interference with the hitting objective. We retain safety and hardware constraint terms, and enable domain randomization and observation noise to consolidate robustness.
\end{itemize}

\subsubsection{Annealed Reward Design across Stages}
We use a modular reward function whose components---locomotion \(r_{\mathrm{loco}}\), hitting \(r_{\mathrm{hit}}\), and global regularization \(r_{\mathrm{global}}\)---are re-weighted across curriculum stages to guide the emergence of coordinated badminton skills. The locomotion reward follows \cite{rudin2022learning} and leverages feet air-time to encode step frequency implicitly.  We follow the hit reward setting in \cite{scirobotics.adu3922} but make several modifications. We group the remaining terms throughout all stages into global regularization, \(r_{\mathrm{global}}\) (action rate, joint limits, collision, energy etc.). 

\textit{S1 — Footwork acquisition.}
Let \(d=\|(\mathbf{p}^{*}_{\mathrm{ee}}-\mathbf{p}_{\mathrm{base}})_{xy}\|\). The main shaping encourages reaching the target region while not necessarily requiring precise approach, being nearby will suffice:
\begin{equation}
r_{\mathrm{track}}=\exp\!\big(-4\,\max(d-0.3,\,0)\big).
\end{equation}
S1 applies standard style shaping for humanoid stable walking—base height and orientation, acceleration regularization, contact-aware footstep terms (feet air-time, touchdown speed, step height, foot posture, no-double-flight), and simple gait symmetry shaping—plus face alignment to the front. 

\textit{S2 — Precision-guided swing generation.}
On top of S1, we lower the weight of regional tracking and introduce a hit-instant reward with large weight (six activations per episode). This reward comes with two terms, hitting precision and racket swinging speed. While \cite{scirobotics.adu3922} splits hitting precision into position and orientation with different scales, we find it to be more accurate to entangle two terms together, coupling position and orientation accuracy \cite{ha2024umi}. Then we introduce another racket speed term to encourage swing. Define the target racket normal \(\mathbf{n}^{*}\) as the \(z\)-axis of target hitting orientation \(\mathbf{q}^{*}_{\mathrm{ee}}\). Let \(\mathbf{v}_{\mathrm{ee}}\) be the end-effector linear velocity; the effective speed component in the direction of target is \(\mathbf{v}_{\mathrm{ee}}\!\cdot \mathbf{n}^{*}\). Let \(e_{ee\_pos}=\|\mathbf{p}_{\mathrm{ee}}-\mathbf{p}^{*}_{\mathrm{ee}}\|\) and \(e_{ee\_ori}=\mathrm{ang\_err_z}(\mathbf{q}_{\mathrm{ee}},\mathbf{q}^{*}_{\mathrm{ee}})\) (defined as the angle between the actual and target racket normals). The hit reward, active only at \(t=t^*\), is
\begin{equation}
\left\{
\begin{aligned}
r_{\mathrm{swing}}
&= \exp\!\Big(-\tfrac{e_{ee\_pos}^2}{\sigma_p}\Big)\,
   \exp\!\Big(-\tfrac{e_{ee\_ori}}{\sigma_r}\Big)
   + 0.3\, r_v, \\
r_v
&= 1-\exp\!\big(-\tfrac{\max(0,\, \mathbf{v}_{\mathrm{ee}}\cdot \mathbf{n}^{*})}{\sigma_v}\big).
\end{aligned}
\right.
\end{equation}
We start with relaxed pose tolerances \(\sigma_p=2.0\), \(\sigma_r=8.0\) to allow a full swing to emerge, and then schedule tightening of position and orientation accuracy to \(\sigma_p=0.1\), \(\sigma_r=1.0\) as training progresses. The racket speed sigma is fixed at \(\sigma_v=3.0\), so a \(5~\mathrm{m/s}\) swing yields \(>80\%\) of the speed term, balancing accuracy with deploy-time stability (incorporating a commanded end-effector velocity during training is feasible). We also add light swing-style regularization to align the racket short-edge (\(y\)) axis with the world \(y\)-axis,
\begin{equation}
r_{y\text{-align}}
= \big(\,\hat{\mathbf y}_{\mathrm{ee}}^\top \hat{\mathbf y}_{\mathrm{world}}\,\big)^{2},
\quad
\hat{\mathbf y}_{\mathrm{ee}} = R(\mathbf{q}_{\mathrm{ee}})\,\mathbf e_y,
\end{equation}
which encourages the policy to adopt a more human-like hitting posture—performing a backswing and then swinging forward along the reverse of that backswing—to produce a complete kinetic chain. We also reward a default holding pose when no shuttle is launched, improving deployment stability and recoverability. Additionally, we introduce collision penalties and strengthen the energy and torque penalties.

\textit{S3 — Task-focused refinement.}
Starting from S2, we anneal away the target-approach reward and several auxiliary gait-shaping terms in \(r_{\mathrm{loco}}\), including foot distance, contact and step symmetry, and step height. This reduces gradient interference between locomotion regularization and the hitting objective. We retain \(r_{\mathrm{hit}}\) and global safety/hardware regularization terms, and enable domain randomization plus observation noise to consolidate robustness.

\subsection{Model-based Hitting Target Generation and Prediction}

Building on the work presented in \cite{scirobotics.adu3922}, we similarly adopted a model-based approach for both generating shuttlecock trajectories required for robot training and predicting hitting positions using physical models.

\subsubsection{Generate Shuttlecock Trajectory for Training}

To generate badminton flight trajectories for training, we employ a physics-based simulation approach. This method adheres to the badminton dynamics model in  \cite{cohen2015physics}, with the shuttlecock's state updated according to the following equation:

\begin{equation}
    m \frac{d\mathbf{v}}{dt} = m\mathbf{g} - m \frac{\|\mathbf{v}\| \mathbf{v}}{L}
    \label{eq:placeholder_label}
\end{equation}

where \(m\) represents the shuttlecock mass, \(v\) denotes its velocity, \(g\) is gravitational acceleration, and \(L\) is the aerodynamic characteristic length, defined as \(L = 2m/\rho SC_D\). Here, \(\rho\) is air density, \(S\) is the cross-sectional area of the shuttlecock, and \(C_D\) is the drag coefficient. The calculated aerodynamic length \(L\) used in our work is 3.4.



Shuttle trajectories are generated and filtered to match realistic hitting conditions. We retain only trajectories whose interception lies within a designated hitting zone \(x \in [-0.8, 0.8]\), \(y \in [-1, 0.2]\), and \(z \in [1.5, 1.6]\) with traversal time \(\ge 0.8~\mathrm{s}\). The \(y\)-range is asymmetric due to the right-handed racket configuration. For each retained trajectory, we store the interception position, orientation, and timing.

\subsubsection{Shuttlecock Trajectory Prediction for Deployment}



During deployment, we use an EKF to estimate the shuttle state online and predict the interception target with the same dynamics model above. When the predicted trajectory first enters the predefined interception box, we record its position and time as the hitting target. The predicted velocity at this time is used to construct the desired racket orientation. The predictor is activated at \(0.36~\mathrm{s}\) after launch once sufficient history is available, and the target is updated in a rolling manner and fed to the controller.


\subsection{Prediction-free variant}
We additionally train a prediction-free policy whose only change from the main pipeline lies in the observation space; the reward function, action space, and annealed curriculum schedule remain unchanged. Specifically, the actor no longer receives the commanded hitting position, orientation, and time $\{\boldsymbol{p}^*_{\mathrm{ee}}, \boldsymbol{q}^*_{\mathrm{ee}}, t^*\}$. Instead, it ingests a sliding window of world-frame shuttle positions comprising the current frame and the previous five frames sampled at 50\,Hz (i.e., \(\{\boldsymbol{p}_{\mathrm{sh}}(t),\boldsymbol{p}_{\mathrm{sh}}(t{-}0.02),\ldots,\boldsymbol{p}_{\mathrm{sh}}(t{-}0.10)\}\)). From this short history, the actor must implicitly infer the intended interception pose and timing. Meanwhile, since each sampled trajectory has a known hit target, the critic and rewards retain privileged access to the ground-truth interception point.

A practical advantage of this variant is that the deployment no longer relies on a hand-tuned predictor (e.g., an EKF plus a parametric aerodynamic model) to deliver $\{\boldsymbol{p}^*_{\mathrm{ee}}, \boldsymbol{q}^*_{\mathrm{ee}}, t^*\}$. Instead, the controller conditions directly on measured shuttle positions, making the overall pipeline more end-to-end. During training, we randomize the aerodynamic parameters per shot, effectively exposing the policy to an “aerodynamic patch” whose exact coefficients need not be known. Intuitively, this mirrors human play: players do not know the drag constants of a specific shuttle a priori but infer landing tendencies from a brief flight segment. We observe that the prediction-free policy achieves hitting performance with only a modest drop relative to the target-known setting, while simplifying deployment by removing the need to tune predictor and physics parameters.

\section{Experiments}

We evaluate our approach through both simulation and real-world experiments. In simulation, we evaluate the effectiveness of our annealed curriculum, conduct ablation studies to investigate the contribution of its stages, and compare it with its prediction-free variant. In the real world, we evaluate the accuracy of shuttle trajectory prediction, the precision of swinging motion, and the overall hitting performance.

\subsection{Experiment Settings}


All simulation studies are performed in Isaac Gym. The evaluation metrics are summarized in Table~\ref{table:evaluation metrics}, where \(\mathcal J\) denotes the set of all joints. \(q\) and \(\tau\) denote the joint angle and joint torque respectively. We define a hit as successful if the impact position error is below 0.10~m and the impact orientation error below 0.2~rad, where 0.10~m is approximately half the racket short-axis width. For ablations, each setting is evaluated over 5 independent runs with different random seeds, and in each run the robot performs 20 consecutive hitting trials with random ball trajectories. All hardware experiments are conducted in a MoCap arena.

\begin{table}
\caption{Evaluation metrics.}
\label{table:evaluation metrics}
\begin{center}
\begin{tabular}{c>{\centering\arraybackslash}m{6cm}}
\hline
\textbf{Metric} & \textbf{Definition}\\
\hline
$e_{ee\_pos}$ & Impact position error.\\
$e_{ee\_ori}$ & Impact orientation error.\\
SR & Hitting success rate.\\
Racket Spd. & Effective racket speed, $\mathbf{v}_{\mathrm{ee}}\!\cdot \mathbf{n}^{*}$.\\
Smooth. & Mean joint acceleration per step, $\frac{1}{|\mathcal J|} \sum_{j\in\mathcal J}|\ddot{q}_j|$.\\
Torque & Mean joint torque per step, $\frac{1}{|\mathcal J|} \sum_{j\in\mathcal J}|\tau_j|$.\\
Energy & Mean joint energy per step, $\frac{1}{|\mathcal J|} \sum_{j\in\mathcal J}|\tau_j \cdot \dot{q}_j|$.\\
\hline
\end{tabular}
\end{center}
\end{table}

\subsection{Simulation Experiments}

\begin{figure}
    \centering
    \includegraphics[width=1\linewidth, height=0.28\textheight]{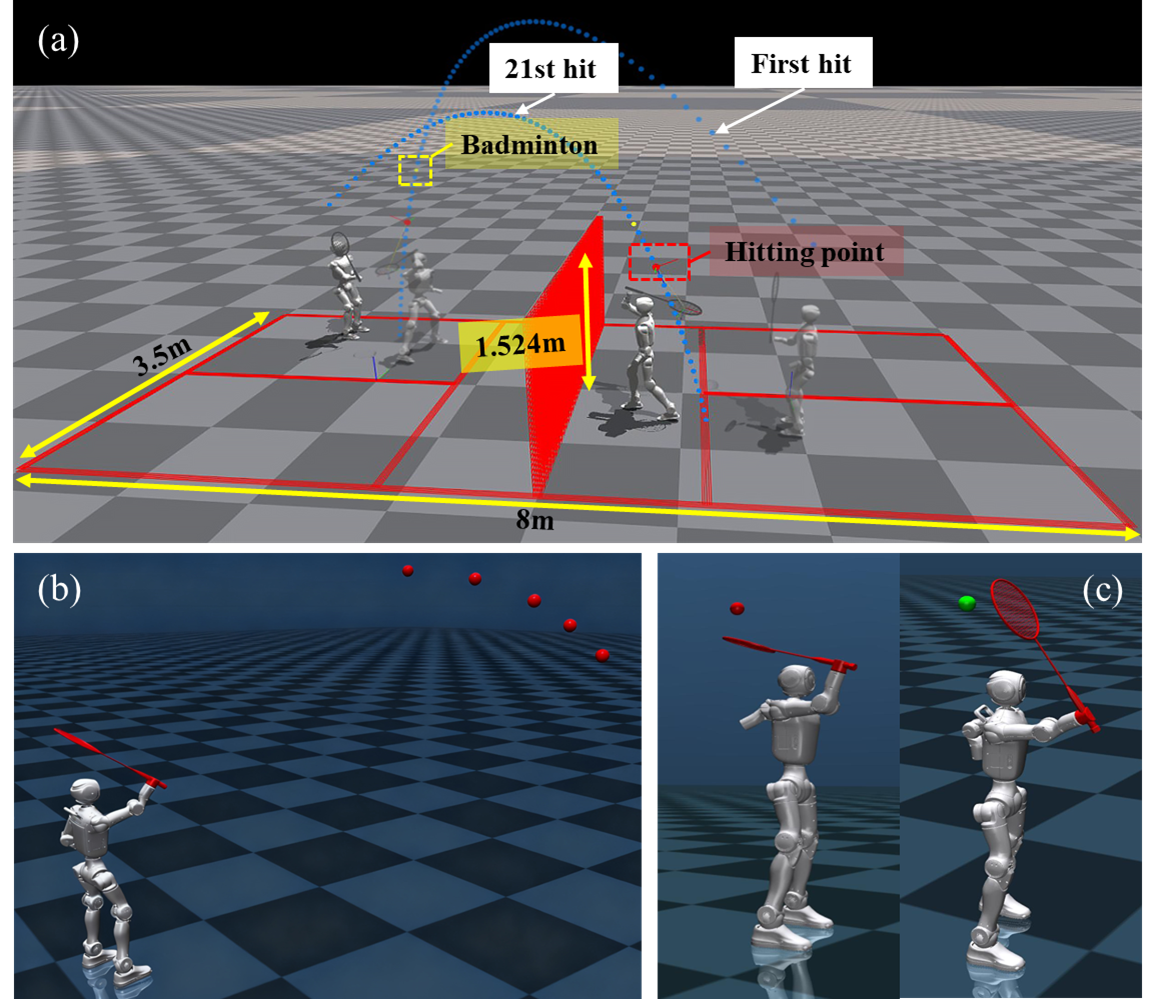}
    \caption{\textbf{Simulation visualizations.} (a) illustrates the multi-rally scenario. (b) presents the prediction-free policy. (c) presents the target-known policy. The red sphere indicates the hitting target, the green sphere confirms a successful hit.}
    \label{fig:figure3}
\end{figure}

\subsubsection{Full Method Performance}

We first evaluate our full method, the target-known policy with annealed curriculum. The quantitative results of all simulation experiments are provided in Table~\ref{table:ablation stats}, where each value is the mean and standard deviation across five seeds. Our full method achieves a 94\% hitting success rate with a racket speed of 4.5~$\mathrm{m/s}$, while producing smooth motions with low energy consumption and joint torque magnitudes, indicating a balance between hitting performance and motion quality.

We further evaluate the policy using two identical humanoids to perform multi-rally play. The evaluation is conducted on a reduced-size court with two humanoids standing on opposite sides, as Fig. \ref{fig:figure3}(a). The first interception target is sampled from the training-style distribution. After a side attempts a hit, the outgoing shuttle velocity is computed from the incoming shuttle state and the racket state at the contact instant, assuming a nearly elastic interaction with racket inertia dominating the shuttle \cite{scirobotics.adu3922}. We then integrate the shuttle dynamics forward until it lands on the opposite half-court and alternate sides. The rally ends if any robot falls, if the shuttle does not clear the net, if it lands out of bounds, or if it contacts the ground. Under these rules, we obtain up to 21 consecutive returns, demonstrating robust long-horizon repositioning, shuttle returning, and posture recovery.

\subsubsection{Annealed Curriculum Ablation}

We further ablate the contribution of each curriculum stage referring to Table~\ref{table:ablation stats}.

First, we analyze the effect of S1, which focuses on acquiring basic locomotion abilities. Skipping S1 makes training prone to failure. The setting w/o S1 terminates before the first hitting trial with no meaningful evaluation. This may be because jointly optimizing locomotion and hitting from scratch creates a large exploration space and causes conflicting gradients between the two objectives. By reducing the entropy coefficient, the setting w/o S1 w/ lower entropy coef. learns a feasible hitting behavior. However, it achieves only a 61\% success rate, a lower racket speed of 3.34~m/s, and less smooth motions, likely due to reduced exploration. These results highlight the importance of learning locomotion before introducing the hitting task.

Then, we investigate the effect of S2, which improves hitting precision gradually. The setting w/o S2 achieves comparable success rate and higher racket speed than the full method, but at the cost of increased joint acceleration and energy consumption, suggesting that training may converge to a local optimum. We expect the accuracy adaptation in S2 to offer more guidance for long-distance interceptions. Thus, we further evaluate the policies on boundary-region hits selected from the training distribution. Both policies degrade under this condition. Without S2, there is a larger reduction in hitting success rate. This demonstrates that S2 is important for handling challenging long-distance interceptions.

Finally, we examine S3, which introduces domain randomization and observation noise while removing redundant lower-body regularization. Despite the harder environment, performance does not regress: joint acceleration and energy consumption decrease by over 30\%, while torque usage drops by 16\%. This suggests that S3 helps break the optimization plateau and leaves room for further policy improvement.

\begin{table*}[h]
\caption{Simulation results.}
\label{table:ablation stats}
\begin{center}
\setlength{\tabcolsep}{2pt}
\begin{tabular}{cccccccc}
\hline
\multirow{2}{*}{\textbf{Setting}} & \multicolumn{4}{c}{\textbf{Hitting performances}} & \multicolumn{3}{c}{\textbf{Regularization terms}}\\
\cmidrule(lr){2-5}
\cmidrule(lr){6-8}
& $e_{ee\_pos}$ ($\mathrm{m}$) $\downarrow$ & $e_{ee\_ori}$ ($\mathrm{rad}$) $\downarrow$ & SR (\%) $\uparrow$ & Racket Spd. ($\mathrm{m/s}$) $\uparrow$
& Smooth. ($\mathrm{rad/s^2}$) $\downarrow$ & Torque ($\mathrm{N \cdot m}$) $\downarrow$ & Energy ($\mathrm{W}$) $\downarrow$\\
\hline
\textit{Full method: }\\
ours & 0.050$\pm$0.004 & 0.072$\pm$0.009 & 94.0$\pm$4.9 & 4.516$\pm$0.097 & 8.343$\pm$0.183 & 3.859$\pm$0.028 & 2.392$\pm$0.087\\
ours w/ boundary hits. & 0.061$\pm$0.013 & 0.092$\pm$0.011 & 85.0$\pm$10.0 & 4.704$\pm$0.241 & 9.849$\pm$0.205 & 4.305$\pm$0.087 & 3.353$\pm$0.163\\
\hline
\textit{Curriculum ablation: }\\
w/o S1 & {---} & {---} & 0.0 & {---} & {---} & {---} & {---}\\
w/o S1 w/ lower entropy coef. & 0.093$\pm$0.014 & 0.137$\pm$0.023 & 61.0$\pm$8.0 & 3.340$\pm$0.072 & 12.779$\pm$0.274 & 4.189$\pm$0.115 & 3.251$\pm$0.185\\
w/o S2 & 0.046$\pm$0.005 & 0.064$\pm$0.014 & 95.0$\pm$5.5 & 5.024$\pm$0.143 & 9.937$\pm$0.404 & 3.806$\pm$0.075 & 2.725$\pm$0.200\\
w/o S2 w/ boundary hits. & 0.063$\pm$0.015 & 0.083$\pm$0.018 & 82.0$\pm$7.5 & 4.882$\pm$0.160 & 11.604$\pm$0.505 & 4.209$\pm$0.137 & 3.674$\pm$0.363\\
w/o S3 & 0.050$\pm$0.007 & 0.064$\pm$0.008 & 94.0$\pm$3.7 & 4.380$\pm$0.094 & 10.956$\pm$0.429 & 4.479$\pm$0.112 & 3.281$\pm$0.238\\
\hline
\textit{Observation ablation: }\\
Prediction-Free & 0.076$\pm$0.003 & 0.083$\pm$0.008 & 73.0$\pm$5.1 & 4.169$\pm$0.332 & 9.817$\pm$0.477 & 3.884$\pm$0.100 & 2.627$\pm$0.255\\
\hline
\end{tabular}
\end{center}
\end{table*}

\subsubsection{Observation Design Ablation}

We then compare two observation designs shown in Fig.~\ref{fig:figure3}(b--c). The target-known policy is evaluated with fixed aerodynamic parameters, while the prediction-free policy with moderately varying aerodynamic characteristic lengths to emulate different shuttles.

As shown in Table~\ref{table:ablation stats}, the prediction-free policy maintains a success rate of 73\%, indicating that explicit target information is not strictly required for hitting. However, the drop in precision and racket speed indicates that explicit target information provides direct and clear guidance for learning more accurate shuttle hitting. We attribute this performance gap mainly to timing ambiguity. Without direct target information, the policy must infer the shuttle trajectory to get impact timing from implicit velocity estimation.

\subsection{Real-World Experiments}


\begin{figure}
    \centering
    \includegraphics[width=1\linewidth]{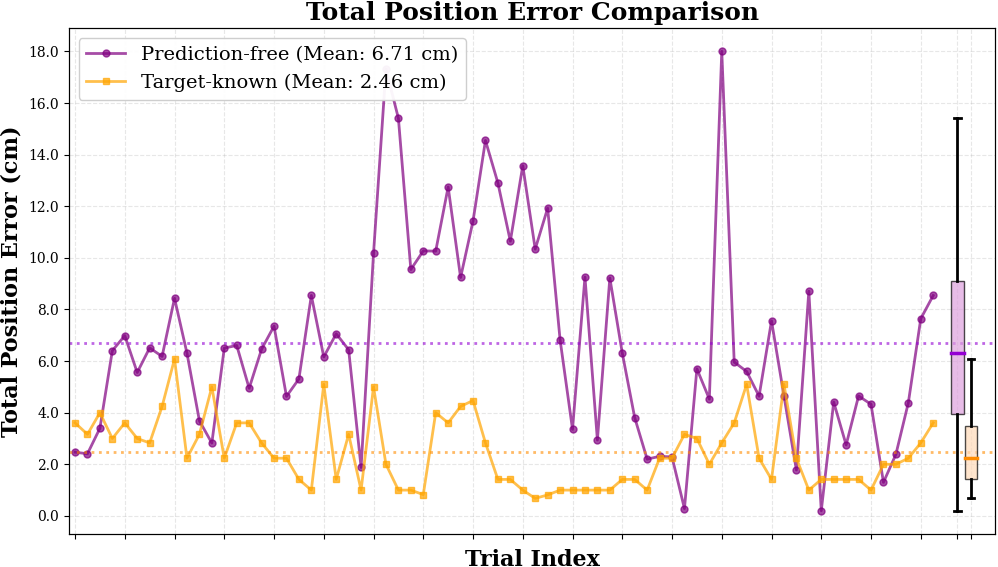}
    \caption{\textbf{Swinging motion precision. }This figure illustrates the Euclidean distance error between the racket face center and the designated hitting position at the moment of impact.}
    \label{fig:figure6}
\end{figure}

\subsubsection{EKF Prediction Accuracy}



We collected 20 real shuttle flight trajectories. A partial segment from each trajectory was provided to the EKF as measurements, and the predicted interception position and timing are compared against ground truth. At 0.6~s before impact, the mean position error is below 100~mm and rapidly converges, reaching 10~mm by 0.3~s before impact. The mean interception timing error is approximately 20~ms at 0.6~s before impact and converges similarly over time. We further perturb the aerodynamic characteristic length by $\pm20\%$ to assess robustness, and observe reduced accuracy but acceptable performance.

\subsubsection{Swinging Motion Precision}

For motion precision evaluation, the robot executes hitting motions without shuttles to isolate the swing behavior. We commanded randomly sampled targets, used MoCap to measure the center position of the racket face, and compared the actual measured interception point against the commanded target position. The corresponding results are presented in Fig. \ref{fig:figure6}.

We conducted 71 swing trials for both policies. The target-known policy reaches a mean hit position error of 2.46 cm, while the prediction-free policy reaches a higher mean error of 6.71 cm. The measured racket speed at impact is typically in the 5–8 m/s range. These results verify that the robot can effectively control the contact point within the radius of the racket face during hitting tasks, demonstrating its accurate racket control for high-precision swing execution.

\subsubsection{Real-World Shuttle Hitting}

\begin{figure}
    \centering
    \includegraphics[width=1.0\linewidth]{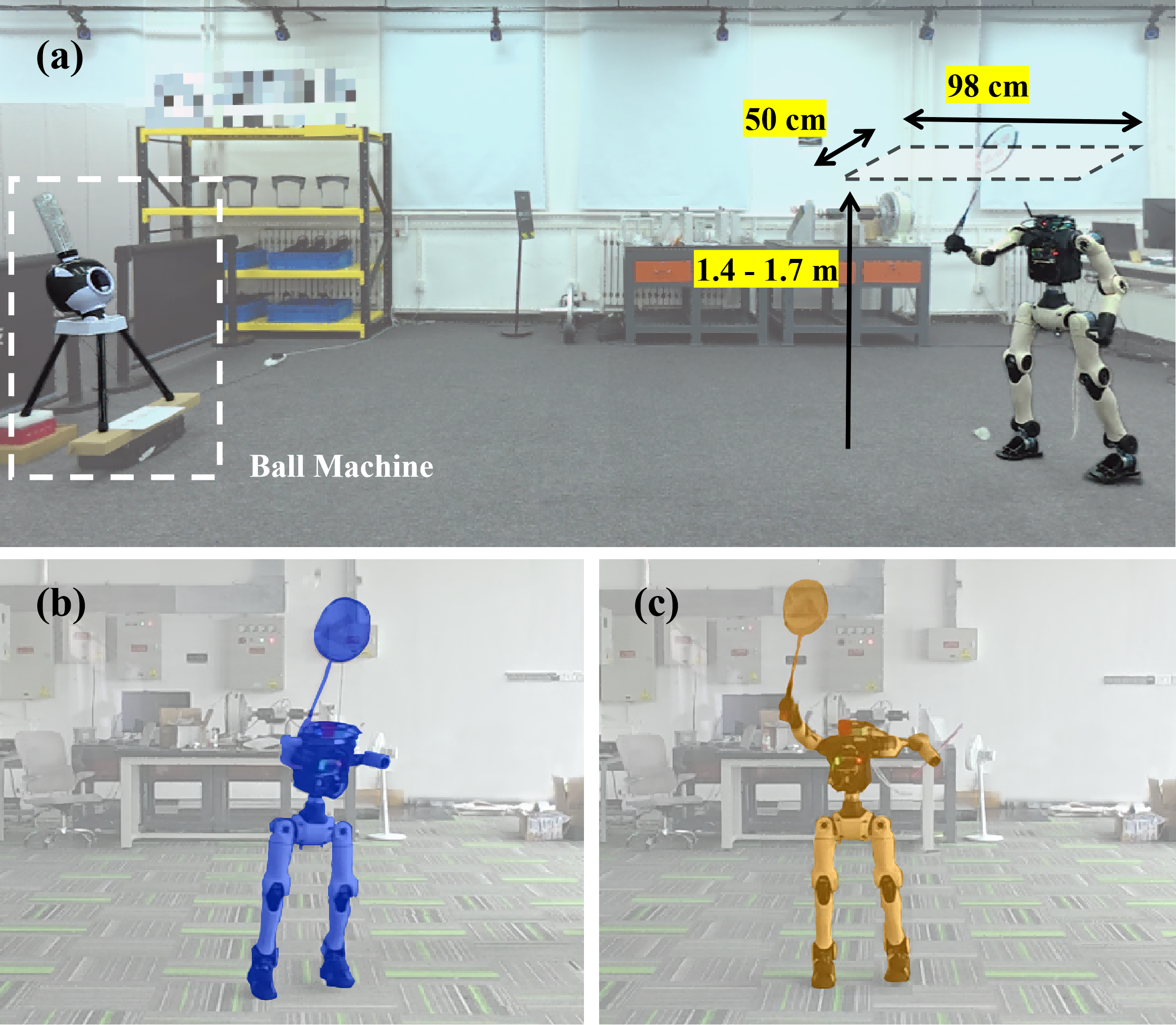}
    \caption{\textbf{Real-world shuttle hitting.} \textbf{(a)} Real-world evaluation with machine-fed shuttles. 
    \textbf{(b--c)} The actual hitting postures at the two opposing boundaries of the interception region.}
    \label{fig:Real-World Shuttle Hitting}
\end{figure}


For real-world shuttle hitting, we first use a ball machine to deliver repeatable shuttles over the tested hitting range shown in Fig.~\ref{fig:Real-World Shuttle Hitting}(a) for quantitative assessment. We collected 46 trials for each policy. For the target-known policy, the robot successfully returns 42 shuttles (success rate 91.3\%), with the remaining 4 hitting the racket frame. For the prediction-free policy, the robot produces 33 valid returns (success rate 71.7\%), with 9 frame hits and 4 misses.

The robot exhibits coordinated whole-body motions, as in Fig.~\ref{fig:Real-World Shuttle Hitting}(b--c). We find that the robot not only intercepts the shuttle but returns it with high quality: the target-known policy achieves outgoing shuttle speeds of up to 19.1~m/s with a mean of 11.1~m/s, whereas the prediction-free policy reaches 18.1~m/s peak speed with an 8.2~m/s mean. The returned shots often follow sharp, steep trajectories with peak heights exceeding 3~m. We further estimate the landing distance of returned shots and obtain an average of roughly 4~m from the interception area.

Given the higher success rate, we use the target-known policy for human--robot rally evaluation, as shown in Fig.~\ref{fig:figure1}. The system can sustain a few rallies with human players with repeatable performance across exchanges, suggesting a path toward real-world badminton interactions.

\section{Discussion}

\subsection{Emergent Whole-Body Coordination}
The  policy exhibits coordinated behaviors beyond those explicitly prescribed by the rewards. For nearby targets, the robot takes short corrective steps before initiating the backswing; for distant or time-critical targets, it begins a long stride and backswing simultaneously. It also occasionally uses tiptoe reaching for high interceptions and re-centers after returns.

Supporting-leg torque analysis further reveals a pronounced peak in the hip-pitch, knee, and ankle-pitch joints approximately 0.1~s before striking, substantially larger than during in-place stepping when no shuttle is incoming. This suggests that the policy learns a pre-strike push-off, using the legs for whole-body impulse generation rather than only locomotion and balance. During fast strokes, the non-racket arm also swings in the opposite direction, which helps counteract overall angular momentum and maintain balance. Together, these observations support the value of unified whole-body control for dynamic badminton strokes.

\subsection{Limitations and Future Work}
Dependence on a MoCap arena restricts operation to instrumented environments and limits deployment flexibility. Moving toward pure vision requires robust shuttle tracking, reliable visual odometry, and policies that can maintain observability by keeping the shuttle within the field of view during aggressive whole-body motion. 

Moreover, the learned striking behavior remains relatively stereotyped. The current system predominantly discovers a single, forehand-like hitting style and does not yet exhibit the diverse repertoire of human strokes.

While we have demonstrated on-court human--robot rallying, sustained long multi-ball rallies remain beyond the current system, largely because the feasible interception area is restricted to a narrow band. Expanding the effective interception workspace by improving whole-body mobility is an important direction for future work.



\section{Conclusions}
We present the first real-world humanoid badminton system driven by a unified whole-body RL policy trained with an annealed curriculum. The curriculum stabilizes early learning with auxiliary locomotion objectives, then progressively removes them to prioritize precise hitting without expert demonstrations. In simulation, two identical humanoids sustain rallies of up to 21 consecutive returns. On hardware, the robot returns machine-served shuttles at outgoing speeds up to 19.1~m/s under sub-second timing and achieves human--robot rallies. A prediction-free variant removes explicit target prediction while retaining comparable performance. Overall, this work is a step toward dynamic, interactive whole-body response tasks on humanoids.

\bibliographystyle{IEEEtran}
\bibliography{references}


\end{document}